\documentclass{article}
\usepackage[utf8]{inputenc}
\usepackage[margin = 1in]{geometry}
\usepackage{authblk}

\usepackage{amsmath, amssymb, amsthm, amsfonts, mathtools, multirow, mathrsfs, tabu, booktabs, lipsum, url, subcaption, graphicx, xcolor, tabu,  authblk, hyperref, blindtext}
\usepackage[normalem]{ulem}
\useunder{\uline}{\ul}{}

\usepackage[english]{babel}
\usepackage[shortlabels]{enumitem}
 \usepackage[ruled,vlined, linesnumbered]{algorithm2e}
\usepackage[capbesideposition=outside,capbesidesep=quad]{floatrow}
\usepackage[framemethod=tikz]{mdframed}
\usepackage{bbm}

\allowdisplaybreaks

\newcommand{\appropto}{\mathrel{\vcenter{
  \offinterlineskip\halign{\hfil$##$\cr
    \propto\cr\noalign{\kern2pt}\sim\cr\noalign{\kern-2pt}}}}}
    
\usepackage[backend=biber,style=ieee,sorting=none,natbib=true]{biblatex} 
\usepackage[utf8]{inputenc}
\usepackage[autostyle=true, threshold=2]{csquotes}

\title{\textbf{Semi-supervised Change Detection of Small Water Bodies Using RGB and Multispectral Images in Peruvian Rainforests}}
\author[1]{Kangning Cui}
\author[2]{Seda~Camalan}
\author[1]{Ruoning Li}
\author[2]{Victor P. Pauca}
\author[2]{Sarra Alqahtani}
\author[2]{Robert J. Plemmons}
\author[3,4]{Miles Silman}
\author[5]{Evan N. Dethier}
\author[5]{David Lutz}
\author[1]{Raymond H. Chan\footnote{Corresponding Author: raymond.chan@cityu.edu.hk}}

\affil[1]{ Department of Mathematics, City University of Hong Kong, Kowloon, Hong Kong}
\affil[2]{ Departments of Computer Science, Wake Forest University, Winston-Salem, NC, USA}
\affil[3]{ Department of Biology, Wake Forest University, Winston-Salem, NC, USA}
\affil[4]{ Center for Energy, Environment, and Sustainability, Wake Forest University, Winston-Salem, NC, USA}
\affil[5]{ Environmental Studies Department, Dartmouth College, Hanover, NH, USA}
\date{}                     
\begin{document}
\topmargin=0mm

\maketitle
\begin{abstract}
Artisanal and Small-scale Gold Mining (ASGM) is an important source of income for many households, but it can have large social and environmental effects, especially in rainforests of developing countries. 
The Sentinel-2 satellites collect multispectral images that can be used for the purpose of detecting changes in water extent and quality which indicates the locations of mining sites.
This work focuses on the recognition of ASGM activities in Peruvian Amazon rainforests. 
We tested several semi-supervised classifiers based on Support Vector Machines (SVMs) to detect the changes of water bodies from 2019 to 2021 in the Madre de Dios region, which is one of the global hotspots of ASGM activities.
Experiments show that SVM-based models can achieve reasonable performance for both RGB (using Cohen's $\kappa$ 0.49) and 6-channel images (using Cohen's $\kappa$ 0.71) with very limited annotations. 
The efficacy of incorporating Lab color space for change detection is analyzed as well.
\end{abstract}


\section{Introduction}

Change detection in remote sensing images involves analyzing the temporal effects of certain phenomena quantitatively using multi-temporal datasets. 
The changes of land coverage provide information in various aspects, including the occurrences of natural disasters, changes in urban areas and agricultural fields, and indications of mining activities, see~\cite{bioucas2013hyperspectral,liu2019review,SedaChange2022}.
Illegal gold mining often impacts the environment by discarding mercury, used by miners to process the ore, directly into streams. 
This can cause a color change of water, which may indicate the location of mining activities~\cite{SedaChange2022,gerson2022amazon}. 
In this work, we have tested several classical and state-of-the-art semi-supervised classifiers and compared the classification results of water change types in Madre de Dios (MDD) regions. 
Semi-supervised approaches require less manual labeling and computational resources during training than supervised classifiers.
In~\cite{SedaChange2022}, SVM with Smoothed Total Variation (SVM-STV) was applied to the difference images of the bi-temporal data. 
SVM-STV was mainly used to classify the water changes which are considered indicators for mining activities. 
Although the data dimension was reduced by taking the difference of the bi-temporal images, the original spectral information of the data was ignored. %
In this work, instead of using difference images, we concatenated the pre-processed bi-temporal datasets of the same region together (also known as image stacking~\cite{liu2019review}) for recognizing mining pond changes.

This article is organized as follows. 
Section 2 introduces the data and methodology of the study, including the study sites, the change categories, and the methods used to process and classify the water changes. 
In Section 3, the results of semi-supervised algorithms are compared and analyzed. 
Section 4 concludes the article and discusses future work.

\begin{figure}[ht]
    \centering
    \includegraphics[width=0.7\textwidth]{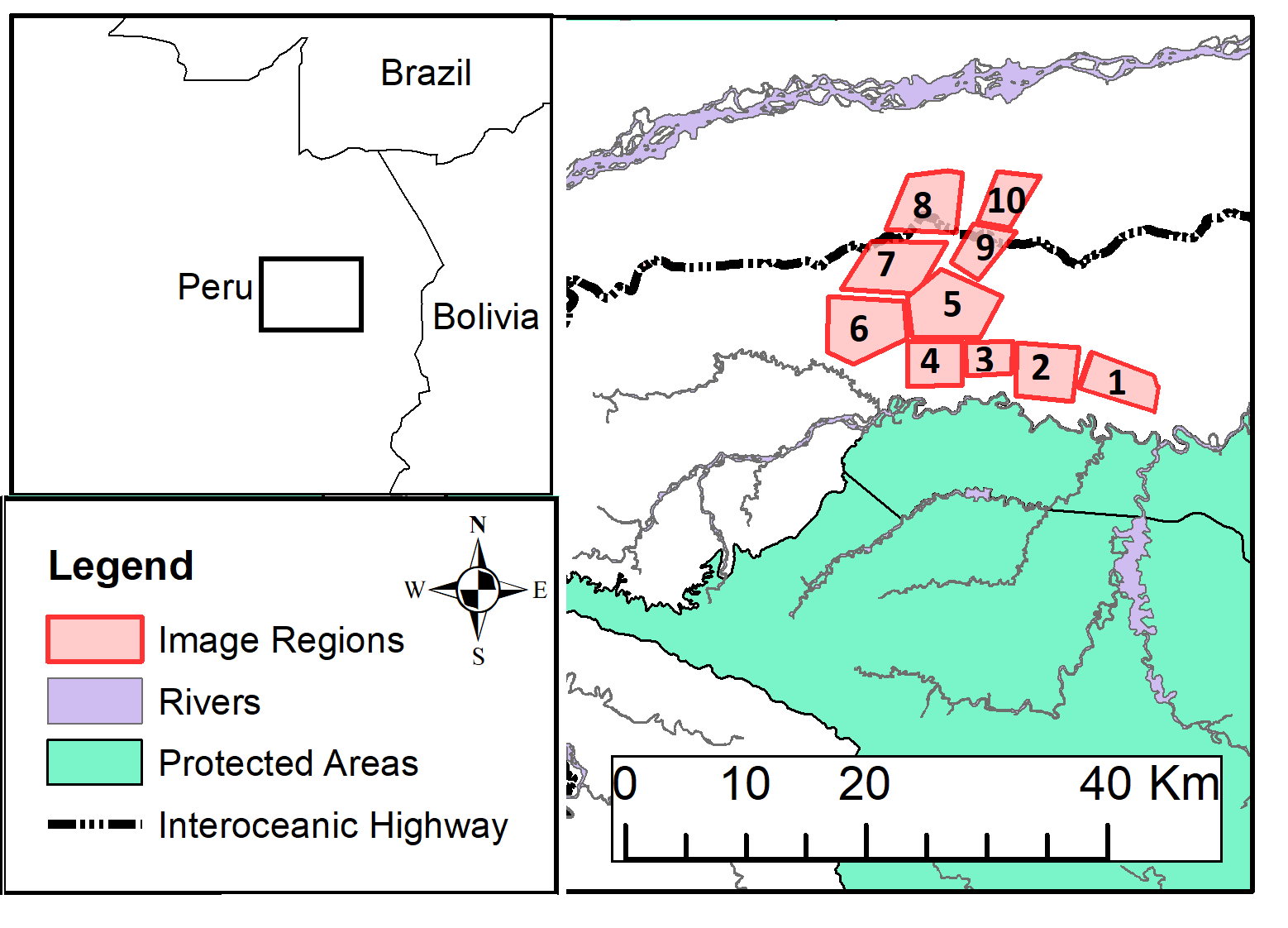}
    \caption{Location of the ASGM study sites in Madre de Dios}
    \label{fig:map}
\end{figure}

\section{Data and Methodology}

\subsection{Study Site and Data Acquisition}

Madre de Dios is a global hotspot of ASGM activity. In this work, we investigated several regions in the La Pampa area, see Figure \ref{fig:map}. 
Ten samples ($\sim 37.7 km^2$ each) of interest in La Pampa are selected according to a gradient of significant mining intensity, techniques, and policy enforcement over the last 15 years.
Therefore, the dates of bi-temporal images are selected to maximize the number of changed pixels and provide more examples for the technique.
To acquire the data, we used the Google Earth Engine (GEE) platform to get the Sentinel-2 images. 
The Sentinel-2 satellite constellation~\cite{drusch2012sentinel} was developed for monitoring variability on land surfaces with 13 multi-spectral channels, ranging from ultra-blue to shortwave infrared. 
Thanks to infrared and shortwave infrared channels, Sentinel-2 data is widely used to assess land surface water change~\cite{cordeiro2021automatic,pahlevan2020seamless}.

\subsection{Change Detection of RGB \& Multispectral Images}

We selected data from two distinct dates (August 18th, 2019, and July 23rd, 2021) which showcase periods in which significant land-use change had occurred in the regions. 
Preferentially selected data are with minimal cloud cover or other atmospheric influences.
Specifically, two different sets of spectral images were composed to create the data. 
One of the sets consists of 3-Channel, Red-Green-Blue images, which have low-cost process and storage. Another set includes 6-Channel, Red-Green-Blue-NIR-SWIR-1-SWIR-2 images, where NIR-SWIR-1-SWIR-2 channels are used to calculate water indices (NDWI and MNDWI) introduced in the following subsection. 

For the purpose of generating data to detect the changes related to ASGM, we defined three pond states of mining: active, transition, and inactive as in \cite{SedaChange2022}. 
\emph{Active state}, in which pond mining was ongoing at the time of collecting the data.
\emph{Transition state}, in which pond mining was recent but not ongoing. 
\emph{Inactive state}, in which mining had ceased more than six months before the imaging time.
Changing categories were then defined as follows~\cite{SedaChange2022}: 
(1) \emph{Decrease}: Change from active to inactive, active to transition, or transition to inactive;
(2) \emph{Increase}: Change from inactive to active, inactive to transition, or transition to active;
(3) \emph{Water Existence/Absence}: change from water to no-water or no-water to water; and, 
(4) \emph{No Change}: no state changes between time periods took place. 
These categories were associated with the intensification, cessation, and the effect of governance of ongoing mining activities~\cite{dethier2019heightened,kahhat2019environmental,caballero2018deforestation}. 

\begin{figure}[ht]
    \centering
    \includegraphics[width=0.75\textwidth]{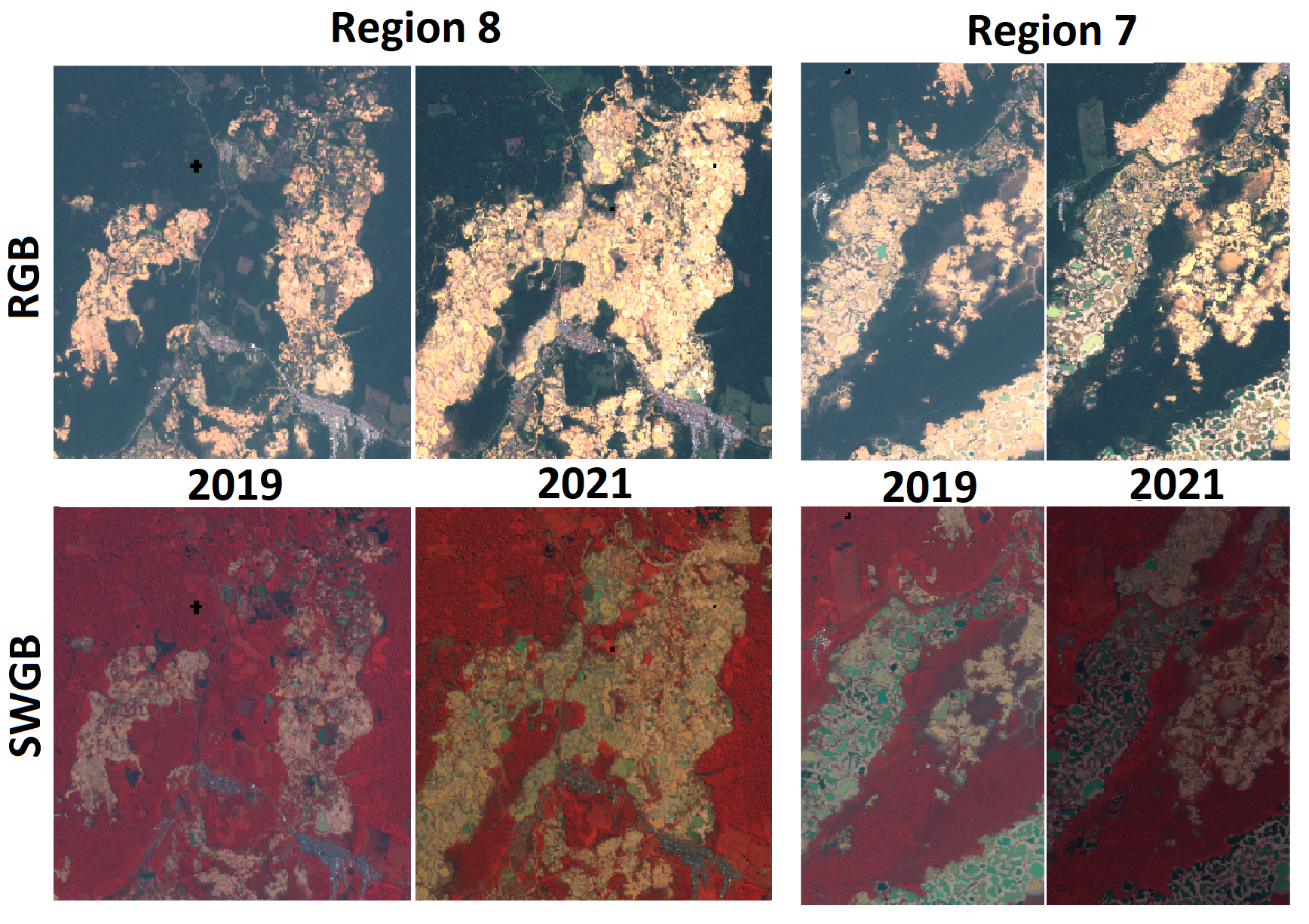}
    \caption{RGB and SWGB Images of $7^{th}$ and $8^{th}$ MDD regions in 2019 and 2021.}
    \label{fig:fig2}
\end{figure}

\subsection{Histogram Matching \& Semi-Manual Labeling}

There is no publicly available dataset for change detection of ASGM activities, and thus we created our own labeled dataset, see~\cite{SedaChange2022}. 
Manual labeling and semi-manual labeling were applied to label ponds and their states. 
To label the ponds manually in relatively small regions, we used RGB and SWGB (SWIR/Green/Blue) images to visualize and distinguish ponds in different colors. 
Because of atmospheric effects, the color shades are different in both RGB and SWGB bi-temporal images (see Fig.\ref{fig:fig2}). 
Thus, we matched the color histograms for each band in RGB and 6-channel images. 

\begin{figure}[ht]
    \centering
    \includegraphics[width=0.6\textwidth]{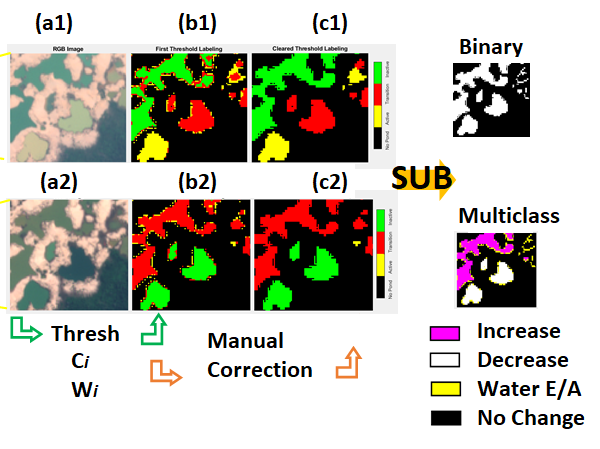}
    \caption{Semi-manual labeling scheme: Thresholding based on color index $Ci$ and Modified Normalized Difference Water Index (MNDWI); manual correction; and subtraction of bi-temporal labels. }
    \label{fig:fig3}
\end{figure}
Manual labeling is time-consuming and labeling based on color is a subjective task and can vary significantly from person to person. 
Therefore, we introduced a semi-manual process based on color histograms and thresholding for consistency labeling. 
Specifically, to label pond states, we used the red and green bands to determine different shades of yellow to green, as sediment reflects highly in the red band (in active ponds). 
Also, photosynthetic material is present (in inactive ponds) and influences the shortwave infrared reflectance.
The color index is defined by $Ci = \frac{(green - red)}{(green + red)}$ to categorize the state of  ponds as shown in our previous work~\cite{SedaChange2022}. 
Distributions of the color index were used to define thresholds (0 and 0.15). 
Furthermore, to accurately classify pond changes, water surfaces are identified before color thresholding to remove nearby vegetation (green) and sand (yellow). 
There have been several studies to detect or segment the surface water body in the defined zones~\cite{donchyts201630,yang2020monthly}. 
There are two formulations of spectral water indices that are the most used methods in the literature~\cite{du2016water}. 
According to the proposed study in ~\cite{donchyts201630}, MNDWI has higher accuracy and lower error than the NDWI water index. Thus, we used the MNDWI spectral water index in this work.

According to the preset thresholds, all pixels in the images are labeled. 
However, many labeled pixels in a pond should be of the same state. 
Fig. \ref{fig:fig3} (b1) and (b2) show that the borders of the ponds are labeled differently than the center of the pond after thresholding. 
We checked the labels of the ponds and manually corrected them as needed.

\subsection{Semi-supervised Change Detection}

Semi-supervised machine learning methods have been widely used in identifying land coverage and detecting changes over time~\cite{SedaChange2022,chan2020two}. 
In this article, we introduce several semi-supervised classifiers based on SVMs.
The basic SVM algorithm finds hyperplanes that maximize margins between classes.
Nonlinear kernel functions enable SVM to project and classify the data in a higher-dimensional space so that classes are linearly separable~\cite{chang2011libsvm}. 
To avoid potential over-fitting, $\nu$-SVM uses a parameter $\nu\in[0,1]$ to control the portion of misclassifications in training processes~\cite{scholkopf2000new}.
The pixel-wise SVM classifiers ignore the spatial information in remotely sensed images, whereas  spatial connectivity is observed in many real scenes.
Incorporating spatial features can improve the performance of classifiers~\cite{chan2020two,camps2006composite,fang2015classification,LiSar2022}.

Several algorithms have been  proposed to include spatial features into the kernel of SVMs. 
In the SVM method with composite kernels (SVM-CK)~\cite{camps2006composite}, spatial and spectral features are extracted and fed into different kernels satisfying Mercer's conditions.
The superpixel-based classification via multiple kernels (SC-MK) method~\cite{fang2015classification} used a graph-based superpixel algorithm to segment the hyperspectral image and extracted feature vectors from superpixels to construct multiple kernels. 
The kernels are  then weighted to form composite kernels and train non-linear SVMs.

Multi-stage SVM-based methods, such as \cite{chan2020two,LiSar2022,LiNsw2022}, make use of spatial and spectral information in different phases separately. The rich spectral information is used to discriminate classes, and the spatial information is utilized to reduce the noise in images. 
In the SVM-STV model~\cite{chan2020two}, a $\nu$-SVM was trained and predicted probability vectors for all pixels in the first stage, and then a smoothed total variation model was applied to reduce the noise in the constructed probability tensor and derive the classification map.
For RGB images, the information of each pixel is limited and the RGB color space is highly correlated.
To further utilize the information of RGB images, lifting was introduced in~\cite{cai2017three}, which transfers the images from RGB to Lab color space and concatenates the RGB and Lab images together.
In our previous work~\cite{SedaChange2022}, the SVM-STV algorithm with the lifting option as a preprocessing step also improves the change detection performance for RGB images substantially.

\section{Numerical Experiments}

We compared and presented the results of $\nu$-SVM~\cite{chang2011libsvm}, SVM-CK~\cite{camps2006composite}, SC-MK~\cite{fang2015classification}, and SVM-STV~\cite{chan2020two} for change detection of ponds in the MDD region. 
For both RGB and 6-channel images, we compared against the variant of SVM-STV with lifting (denoted as SVM-STV$'$) as well~\cite{SedaChange2022}.
As the MDD datasets are imbalanced with a dominating class ``no change'', instead of using accuracy, we used Cohen's kappa coefficient $\kappa$, Jaccard index $J$, and $F1$-score to quantify the results~\cite{SedaChange2022}. 
We ran ten trials with different randomly selected training sets to remove the effect of stochasticity.

\begin{figure}[h]
    \centering
    \includegraphics[width=0.5\textwidth]{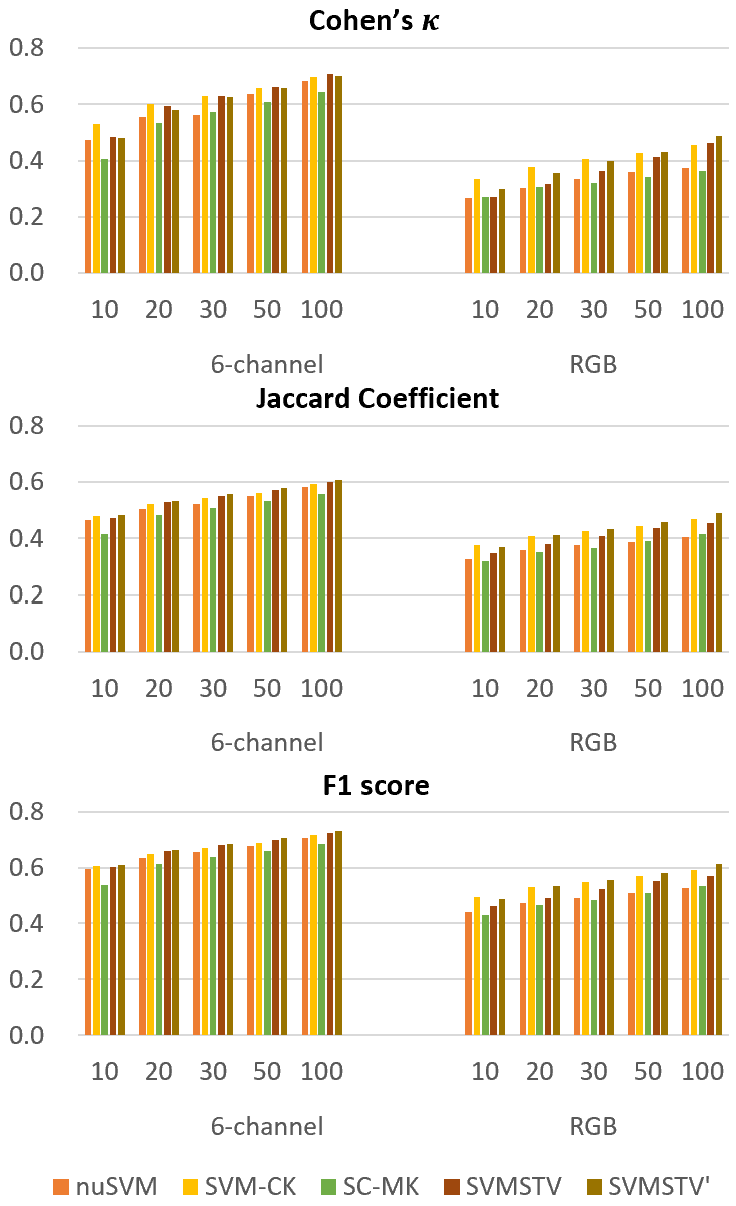}
    \caption{We provide the average performance of the semi-supervised methods using RGB and 6-channel multispectral images with varying training size. The $x$-axes and $y$-axes represent the number of labels per class for training and the corresponding average performance on 10 regions, resp.}
    \label{fig:numerics}
\end{figure}

\begin{figure}[ht]
    \centering
    \includegraphics[width = \textwidth]{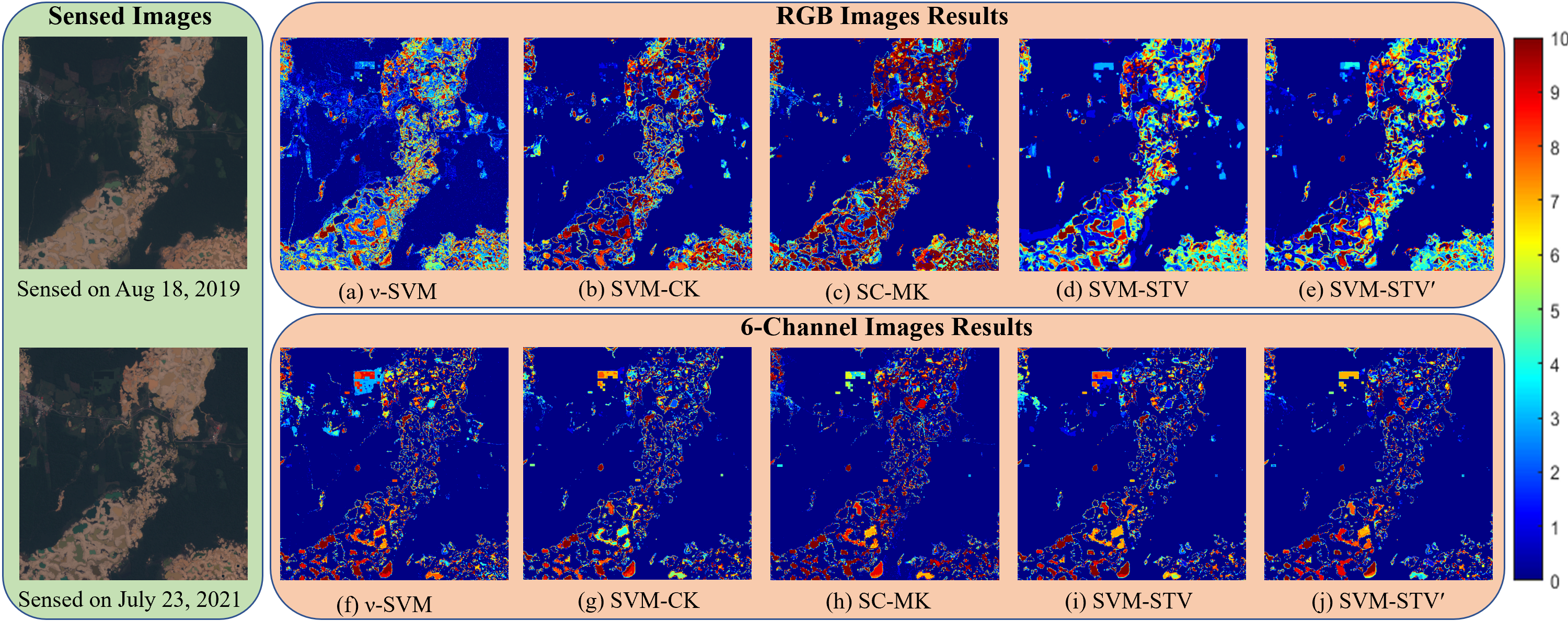}
    \caption{Comparison of classification results using semi-supervised methods on the $9^{th}$ MDD region. The sensed bi-temporal images are shown in the left. The two figures in the right show the heat maps of misclassifications of RGB and 6-channel results, where the colors of pixels indicates the number of false classifications in 10 trials (see the colorbar).} 
    \label{fig:results}
\end{figure}

Fig. \ref{fig:numerics} shows the average quantitative performance across the ten regions in MDD. 
The results improve monotonically with increasing labels.
Comparing 6-channel results with RGB results, the portion of falsely classified pixels is highly reduced because infrared channels in the 6-channel images provide essential information to discriminate water changes. 
SVM-CK outperforms other methods in terms of $\kappa$ when the number of labels/class $\leq30$. 
SVM-STV or its variant outperforms other methods in the three metrics in other cases.
Table \ref{tab:results} shows the classification performance of compared SVM-based methods when No. labels/class $=100$. 
It shows that SVM-STV$'$ outperforms SVM-STV by 2-4\% in terms of three metrics using RGB images, which indicates that the lifting option provides complementary information that aids the classification of SVM-STV. 
Lifting is not recommended for the 6-channel images since the improvement is negligible and the computational time increases.

\begin{table}[h]
\centering
\resizebox{\textwidth}{!}{%
\begin{tabular}{|c|llllll|llllll|}
\hline
                    & \multicolumn{6}{c|}{\textbf{RGB Images}}                                                                                                                                                                            & \multicolumn{6}{c|}{\textbf{6-Channel Images}}                                                                                                                                                                      \\ \cline{2-13} 
                    & \multicolumn{2}{c}{\textbf{Kappa}}                                   & \multicolumn{2}{c}{\textbf{Jaccard}}                                 & \multicolumn{2}{c|}{\textbf{F1}}                                      & \multicolumn{2}{c}{\textbf{Kappa}}                                   & \multicolumn{2}{c}{\textbf{Jaccard}}                                 & \multicolumn{2}{c|}{\textbf{F1}}                                      \\
                    & \multicolumn{1}{c}{\textbf{Avg}} & \multicolumn{1}{c}{\textbf{Best}} & \multicolumn{1}{c}{\textbf{Avg}} & \multicolumn{1}{c}{\textbf{Best}} & \multicolumn{1}{c}{\textbf{Avg}} & \multicolumn{1}{c|}{\textbf{Best}} & \multicolumn{1}{c}{\textbf{Avg}} & \multicolumn{1}{c}{\textbf{Best}} & \multicolumn{1}{c}{\textbf{Avg}} & \multicolumn{1}{c}{\textbf{Best}} & \multicolumn{1}{c}{\textbf{Avg}} & \multicolumn{1}{c|}{\textbf{Best}} \\ \hline
\textbf{$\nu$-SVM}  & 0.3743                           & 0.4383                            & 0.4050                           & 0.8283                            & 0.5260                           & 0.9059                             & 0.6831                           & 0.7125                            & 0.5817                           & 0.9262                            & 0.7047                           & 0.9616                             \\
\textbf{SVM-CK}     & 0.4552                           & 0.5084                            & {\ul 0.4675}                     & 0.8534                            & {\ul 0.5907}                     & 0.9207                             & 0.6961                           & 0.7283                            & 0.5941                           & {\ul 0.9325}                      & 0.7153                           & 0.9650                             \\
\textbf{SC-MK}      & 0.3613                           & 0.3954                            & 0.4145                           & 0.7823                            & 0.5353                           & \multicolumn{1}{c|}{0.8775}        & 0.6440                           & 0.6719                            & 0.5589                           & 0.9113                            & 0.6833                           & 0.9535                             \\
\textbf{SVM-STV}    & {\ul 0.4639}                     & {\ul 0.5552}                      & 0.4556                           & \textbf{0.9102}                   & 0.5691                           & \textbf{0.9529}                    & \textbf{0.7072}                  & \textbf{0.7426}                   & {\ul 0.6020}                     & \textbf{0.9354}                   & 0.7223                           & \textbf{0.9666}                    \\
\textbf{SVM-STV$'$} & \textbf{0.4871}                  & \textbf{0.5685}                   & \textbf{0.4919}                  & {\ul 0.8983}                      & \textbf{0.6134}                  & {\ul 0.9463}                       & {\ul 0.6998}                     & {\ul 0.7367}                      & \textbf{0.6078}                  & 0.9331                            & \textbf{0.7304}                  & {\ul 0.9653}                       \\ \hline
\end{tabular}%
}
\caption{The best and the average performance of semi-supervised methods on 10 MDD regions when the number of labels/class $=100$ for training. The bold and underlined values represent the best and second-best performance, resp.}
\label{tab:results}
\end{table}

Fig. \ref{fig:results} visualizes the heatmaps of misclassifications of these methods on the $9^{th}$ MDD region in Fig. \ref{fig:map} as an example. 
Incorporating spatial information improves the classification at the upper-left part most effectively for the 6-channel images, where the pond change has not occurred. 
The spatial information helps with ironing out the isolated classification noise.
For RGB images, SVM-STV and SVM-STV$'$ predict the pixels located at the middle and lower-right corner of the region better than their competitors, where the pond changes occur extensively.
This is because the spectral information of RGB images is not sufficient for accurate classification, even for neural networks~\cite{SedaChange2022}. 
The smoothed total variation term reduces the spatial noise in the probability maps produced by $\nu$-SVM and thus improves the $\nu$-SVM result. 
SVM-CK upweights spatial information into composite kernels, so it has comparable performance as SVM-STV$'$ when spectral information is very limited.
However, since SC-MK extracts superpixels in the first stage and then extracts spatial features, the quality of superpixels directly impacts the quality of its classification result.

\section{Conclusions}

In this work, we tested several SVM-based classifiers for recognizing water pond changes in Peruvian rainforests, specifically in Madre de Dios. 
Band selection, histogram matching, semi-manual labeling, lifting, and image stacking are performed before classifying the types of water changes. 
Semi-supervised classifiers are then trained with a few expert labels and applied to predict the changes using preprocessed RGB and 6-channel bi-temporal datasets.
These SVM-based methods achieve reasonable results on the ten samples from MDD.
Spatial regularity is observed in the MDD datasets, and incorporating spatial information improves the performance of both RGB and 6-channel images.

Observing that the best performance in the ten trials is much better than the average performance of the ten trials with randomly initialized training labels, a practical-based semi-supervised way called active learning is planned as  future work. 
Instead of selecting labels arbitrarily, active learning methods choose the labels of the most informative pixels for training~\cite{murphy2018unsupervised,ADVIS}.
Unsupervised diffusion-based clustering algorithms will also be tested on ASGM datasets as well~\cite{DVIS,cui2022unsupervised}.

\printbibliography

\end{document}